\documentclass[10pt,twocolumn,letterpaper]{article}

\usepackage[pagenumbers]{cvpr} 

\usepackage{graphicx}
\usepackage{amsmath}
\usepackage{amssymb}
\usepackage{booktabs}
\usepackage{float}
\usepackage{mathtools}
\usepackage{algorithm}
\usepackage{algorithmic}
\usepackage{enumitem}
\usepackage{lipsum}

\usepackage[table]{xcolor}
\usepackage{ragged2e} 
\usepackage{booktabs,makecell, multirow, tabularx}
\usepackage{stfloats}
\usepackage[symbol]{footmisc}

\usepackage[pagebackref,breaklinks,colorlinks]{hyperref}

\usepackage[capitalize]{cleveref}
\crefname{section}{Sec.}{Secs.}
\Crefname{section}{Section}{Sections}
\Crefname{table}{Table}{Tables}
\crefname{table}{Tab.}{Tabs.}

\makeatletter
\def\thanks#1{\protected@xdef\@thanks{\@thanks
        \protect\footnotetext{#1}}}
\makeatother

\begin{document}

\title{MagicFusion: Boosting Text-to-Image Generation Performance \\ by Fusing Diffusion Models}

\vspace{-16mm}

\author{
  Jing Zhao$^{1*}$, Heliang Zheng$^{2}$, Chaoyue Wang$^{2}$, Long Lan$^{1}$, Wenjing Yang$^{1^\dagger}$\\
  \normalsize National University of Defense Technology$^1$,\\
  \normalsize JD Explore Academy$^2$ \\
}

\twocolumn[{%
\renewcommand\twocolumn[2][]{#1}%
 \maketitle%
\centering \centering
\includegraphics[width=\textwidth]{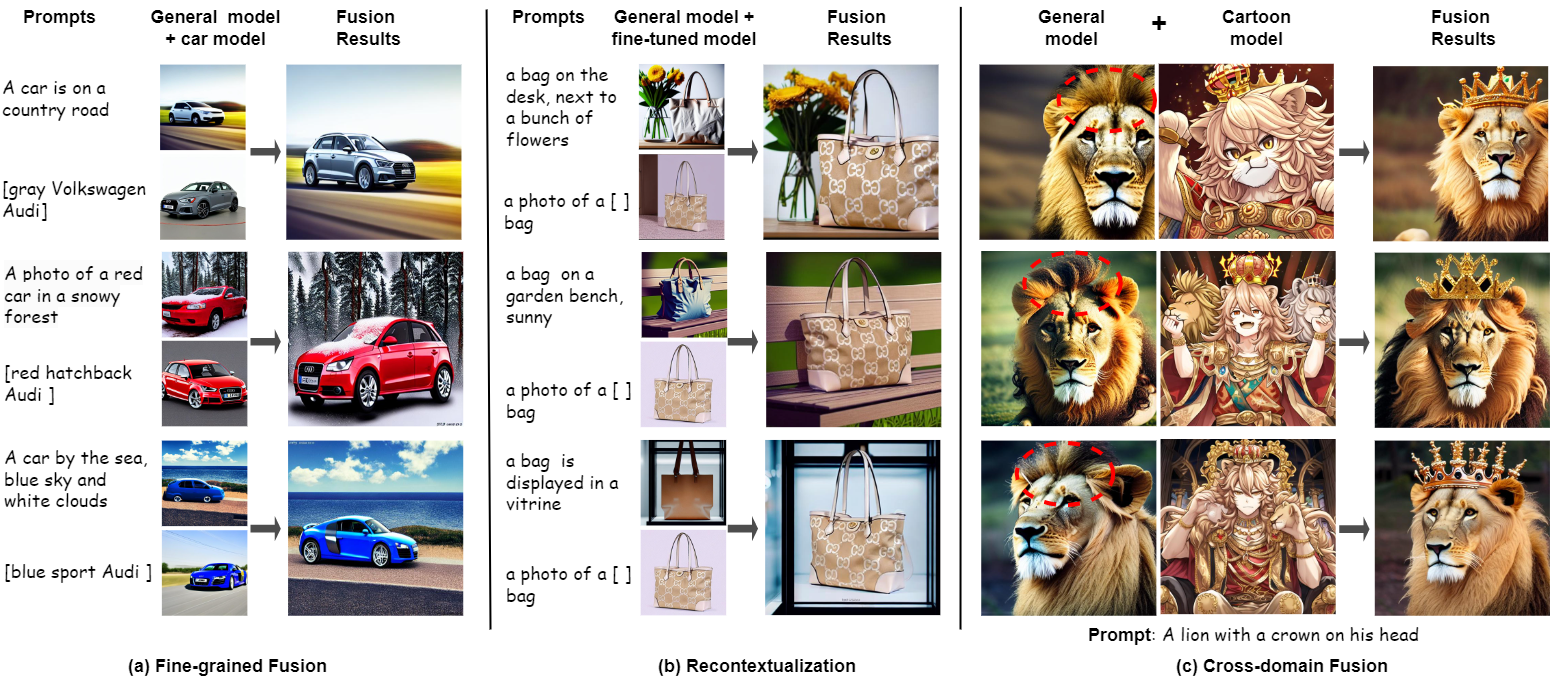}
\captionof{figure}{``MagicFusion''. Given two diffusion models, our method can preserve the strengths of each individual model. (a) A general model~\cite{rombach2022high} + a fine-grained car model to achieve fine-grained generation with complex scenes. (b) a general model + a DreamBooth model~\cite{ruiz2022dreambooth} to recontextualize specific objects with well-preserved details. (c) a general model + a cartoon model to generate creative scenes with photorealistic fidelity.}
\vspace*{0.5cm}
\label{fig:teaser}}]

\footnotetext[1]{Work done during an internship at JD Explore Academy.} 
\footnotetext[2]{Corresponding authors.} 

\begin{abstract}

The advent of open-source AI communities has produced a cornucopia of powerful text-guided diffusion models that are trained on various datasets. While few explorations have been conducted on ensembling such models to combine their strengths. In this work, we propose a simple yet effective method called Saliency-aware Noise Blending (SNB) that can empower the fused text-guided diffusion models to achieve more controllable generation. Specifically, we experimentally find that the responses of classifier-free guidance are highly related to the saliency of generated images. Thus we propose to trust different models in their areas of expertise by blending the predicted noises of two diffusion models in a saliency-aware manner. SNB is training-free and can be completed within a DDIM sampling process. Additionally, it can automatically align the semantics of two noise spaces without requiring additional annotations such as masks. Extensive experiments show  the impressive effectiveness of SNB in various applications. Project page is available at \href{https://magicfusion.github.io/}{https://magicfusion.github.io/}.

\end{abstract}

\section{Introduction}

In recent years, significant progress has been made in image generation ~\cite{db_56, dhariwal2021diffusion, db_52, db_51, rombach2022high, nichol2021glide, db_56, chang2023muse} thanks to breakthroughs in diffusion models ~\cite{db_58, db_56, song2020denoising} and large-scale training~\cite{db_52, nichol2021glide, db_56, chang2023muse}, as well as the contributions of open-source AI communities. Pre-trained large models have become an invaluable resource in this field. One of the most exciting developments has been text-guided diffusion models~\cite{db_52, nichol2021glide, db_56, rombach2022high}. A wide range of powerful text-guided diffusion models trained on various datasets has been publicly released. For example, general models (e.g., stable diffusion v1-4, v1-5, v2-1, etc. ~\cite{rombach2022high})  trained on large-scale multimodal datasets like LAION 5B~\cite{schuhmann2022laion}, as well as more specialized models (e.g., Anything-v3) trained on cartoon and anime datasets or fine-grained categories such as cars~\cite{yang2015large}. There are even fine-tuned models designed for specific objects~\cite{ruiz2022dreambooth}. The vast amounts of data and computational cost have enabled these models to achieve impressive capabilities in various fields. However, few explorations have been conducted on ensembling such models to combine their strengths.

Some work propose to add special symbols or signature phrases when fine-tuning models on new datasets~\cite{modi}. This approach enables the model to generate novel image distributions while retaining its ability to generate from the original data distribution. However, there has been limited discussion on how to effectively combine the generation capabilities of these two distributions. One intuitive method for integrating the capabilities of two models involves taking a weighted average of their predicted noises~\cite{merge}. However, such kind of fusions often fail to fully preserve the strengths of each individual model. Blended diffusion ~\cite{avrahami2022blended} proposes to spatially blend a noisy image and a predicted one, which has been explored in image editing tasks. However, this typically requires specifying a mask to edit particular objects, and few discussions have been conducted to blend the noises of two diffusion models.

In this work, we propose a simple yet effective method called Saliency-aware Noise Blending (SNB) that can empower the fused text-guided diffusion models to achieve more controllable generation. Specifically, we integrate two diffusion models by spatially blending the predicted noises. Our insight is to trust different models in their areas of expertise, thus the strengths of each individual model can be preserved. To obtain diffusion models' areas of expertise, we revisit the classifier-free guidance~\cite{ho2022classifier}, which is widely adopted in text-guided diffusion models to enhance the difference between a given text and a null text in the predicted noise space. We experimentally find that the responses of classifier-free guidance are highly related to the saliency of generated images. To this end, we propose Saliency-aware Noise Blending that blends the predicted noises of two diffusion models based on their responses of classifier-free guidance. 

SNB is training-free and can be completed within a DDIM sampling ~\cite{song2020denoising} process. Additionally, it can automatically align the semantics of two noise spaces without requiring additional annotations such as masks. Our main contributions can be summarised as follows:
\begin{itemize}
\item We propose to fuse two well-trained diffusion models to achieve more powerful image generation, which is a novel and valuable topic.
\item We propose a simple yet effective Saliency-aware Noise Blending method for text-guided diffusion models fusion, which can preserve the strengths of each individual model.
\item We conduct extensive experiments on three challenging applications (\textit{i.e.}, a general model + a cartoon model, a fine-grained car model, and a DreamBooth~\cite{ruiz2022dreambooth} model), and prove that SNB can significantly empower pre-trained diffusion models.
\end{itemize}
The remainder of the paper is organized as follows. We describe related work in Section~\ref{rw}, and introduce our proposed SNB method in Section~\ref{method}. An evaluation on three applications and comparisons are presented in Section~\ref{exp}, followed by conclusions in Section~\ref{con}.

\section{Related Works}
\label{rw}
\begin{figure*}[!h] 
\begin{center}
\includegraphics[width=0.9\linewidth]{ 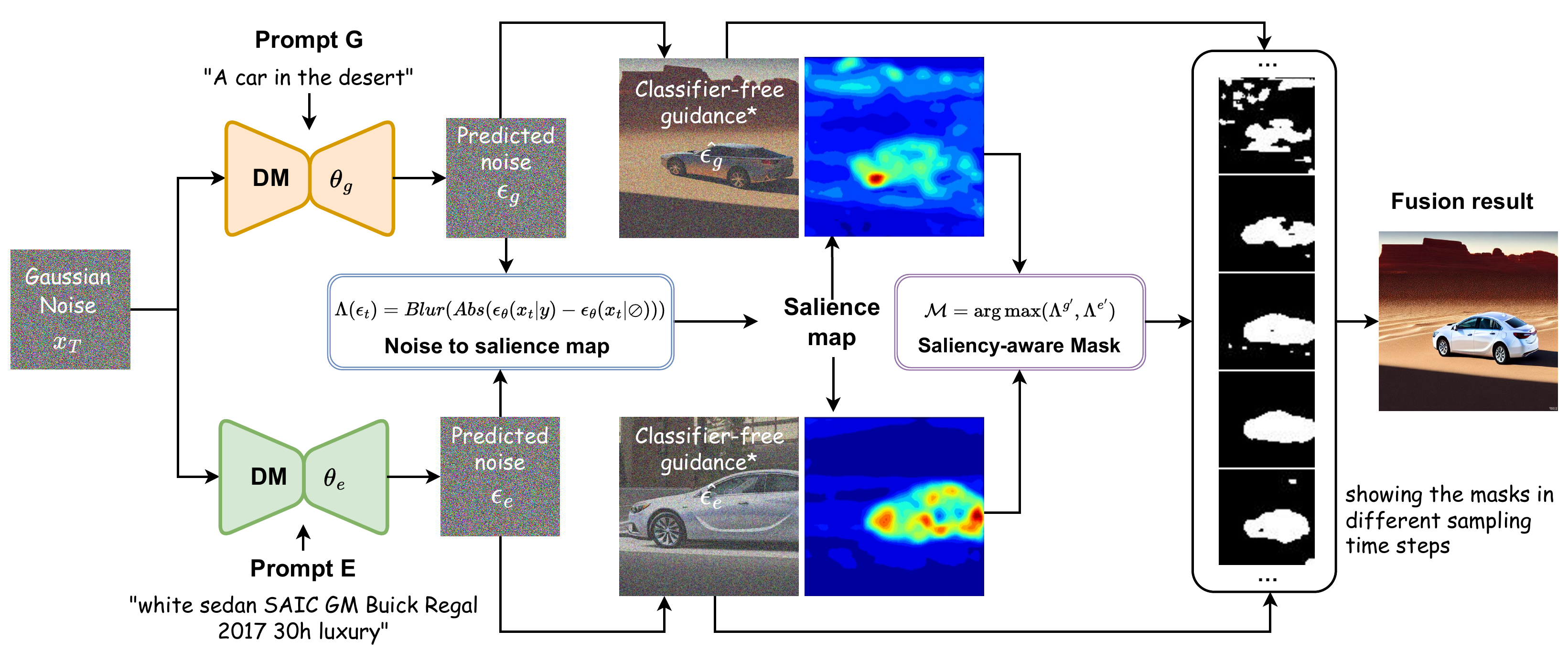}
\end{center}
   \caption{An overview of our Saiency-aware Noise Blending. Given two diffusion models, we first design a ``Noise to salience map'' module to  obtain salience maps. After that, we can generate saliency-aware masks based on the salience maps. Finally, we blend the diffusion models in the noise space according to the mask. (*) $\hat\epsilon_g$ and $\hat\epsilon_e$ are noises instead of noisy images, and we add the image here just for visualization.}
\label{fig:frame}
\end{figure*}

\subsection{Text-to-image synthesis}
Text-guided image generation plays a significant role in image generation~\cite{db_16,db_24,db_64,db_36,db_37,db_47,db_48,db_52,db_67,long2021blind}. Previous works mainly focus on GAN-based ~\cite{db_22} models and small-scale image-text datasets~\cite{zhu2019dm,tao2020df,TaoXu2017AttnGANFT,zhang2021cross,ye2021improving}. Since Transformer-based autoregressive models ~\cite{esser2021taming, db_52} were proposed, more and more attentions have been attracted to large-scale training. The emergence of denoising diffusion models is another milestone, which significantly boosting the generation fidelity ~\cite{db_51,saharia2022photorealistic,nichol2021glide,rombach2022high, chen2021flocking}. Notably, Stable Diffusion ~\cite{rombach2022high} is publicly released, enabling a large number of variants that fine-tuned on different datasets. The vast amounts of data and computational cost have enabled these models to achieve impressive capabilities in various fields. However, few explorations have been conducted on ensembling such models to combine their strengths. 

\subsection{Model Ensembling}
Model ensembling is a powerful technique to distill the knowledge of multiple models and boost the performance, which is widely obtained in image understanding tasks, i.e., classification problems~\cite{zhao2005survey,rokach2010ensemble,gopika2014analysis,yang2010review}, regression problems ~\cite{mendes2012ensemble,ren2015ensemble} and clustering~\cite{vega2011survey}. While such methods are hard to be adapted to generative models due to the large and complex image pixel space. Vision-aided GAN~\cite{kumari2022ensembling} proposes to ensembling pre-trained vision models as a loss to guide the optimization of a generator. eDiff-I~\cite{balaji2022ediffi} proposes to ensemble different denoisers in different timesteps to improve the overall performance of image generation. Composable diffusion model~\cite{liu2022compositional} propose to factorize the compositional generation problem, using different diffusion models to capture different subsets of a compositional specification. These diffusion models are then explicitly composed together to generate an image. In this work, we propose to ensemble different pre-trained diffusion models in a novel dimension, \textit{i.e.}, spatial, which can be applied to various scenarios.

\section{Method}
\label{method}

In this section, we introduce our proposed saliency-aware noise blending. Specifically, we first review the widely obtained classifier-free guidance, after that, we revisit such guidance and experimentally find that the classifier-free guidance is secretly a saliency indicator. Based on the salience map, we can obtain a saliency-aware mask, which is further used to guide the blending the noise of two diffusion models. Figure~\ref{fig:frame} show the whole pipeline, and more details can be found in the following.

\subsection{Preliminaries}

Given a pre-trained text-guided diffusion model, we can generate images by a DDPM/DDIM sampling process, which progressively convert a Gaussain noise into an image for $T$ timesteps. Take the DDPM sampling for example, a denoising step can be denoted as:
\begin{equation}\label{xt_1}
    x_{t-1}=\sqrt{\bar{\alpha}_{t-1}}(\frac{x_t-\sqrt{1-\bar{\alpha} _t}\hat{\epsilon} }{\sqrt{\bar{\alpha}_t} } )+\sqrt{1-\bar{\alpha }_{t-1} }\hat{\epsilon}, 
\end{equation}
where $t$ indicates the timestep, $x$ is the noisy image, $\bar{\alpha}$ is related to a pre-defined variance schedule, and $\hat{\epsilon}$ is the predicted noise. The predicted noise can be re-modulated by classifier-free guidance~\cite{ho2022classifier}, which is designed to extrapolate the output of the model in the direction of $\epsilon_\theta(x_t|c)$ and away from $\epsilon_\theta(x_t|\oslash)$ as follows:
\begin{equation}\label{scala}
    \hat{\epsilon} = \epsilon_{\theta} (x_t |\oslash ) + s \cdot (\epsilon_{\theta } (x_t|c) - \epsilon_{ \theta } (x_t|\oslash )),
\end{equation}
where $\epsilon_{\theta}$ is the pretrained model, $c$ is the text condition, $\oslash$ is a null text, $s$ is the guidance weight and increasing $s>1$ strengthens the effect of guidance.

\subsection{Noise to Salience Map}
We experimentally find that the classifier-free guidance introduced above is secretly a saliency indicator. Specifically, $\epsilon_{\theta } (x_t|c) - \epsilon_{ \theta } (x_t|\oslash )$ in Eqn.~\ref{scala} indicates the difference between a conditional prediction and a unconditional prediction, thus the objects and scenes appeared in the text condition would be emphasized with large value, especially when we adopt a large guidance scale $s$ (\textit{e.g.}, 10-100). We visualize the re-modulated noises of Eqn.~\ref{scala} and find that the important region dose have high responses. To this end, we propose to obtain salience map by the following operation:

\begin{equation}
    \label{saliency}
    \Lambda(\epsilon_t)  = Blur(Abs(\epsilon_{\theta}(x_t | c)-\epsilon_{\theta}(x_t | \oslash)))
\end{equation}
where $\Lambda(\epsilon_t)$ represents the salience map, $Abs(\cdot)$ calculates the absolute value of the input variables, and $Blur(\cdot)$ is used to smooth the high-frequency noise, which can eliminate local interference responses and leverage the coherence of adjacent regions.

\begin{algorithm}[t]
	\caption{Saliency-aware Noise Blending} 
        \label{HPLA}
	\renewcommand{\algorithmicrequire}{\textbf{Input:}}
	\renewcommand{\algorithmicensure}{\textbf{Output:}}
	\begin{algorithmic}[1]
		\REQUIRE Two pre-trained models $\epsilon_{\theta_g}$and  $\epsilon_{\theta_e}$, along with two prompts, $c_g$ and $c_e$. gradient scale $s$ in Eq.\eqref{scala}. Hyperparameters $k^a$ and $k^b$ in Eq. \eqref{lambda_}. 
		\ENSURE The fused image $x_0$
            \STATE $x_T \sim \mathcal{N}(0,I)$
		\FOR {$t$ from $T$ to $0$ }
		\STATE $\epsilon_g = \epsilon_{\theta_g} (x_t |c_g)$,$\epsilon_e = \epsilon_{\theta_e} (x_t | c_e)$
            \STATE get $\Lambda^g(\epsilon_g)$ and $\Lambda^e(\epsilon_e)$ according to Eq.\eqref{saliency}.
            \STATE get $\Lambda^{g'}(\epsilon_g)$ and  $\Lambda^{e'}(\epsilon_e)$ via Eq. \eqref{lambda_}.
            \STATE $\mathcal{M}=\arg\max(\Lambda^{a'},\Lambda^{b'})$
            \STATE get classifier-free guidance  $\hat{\epsilon}_g$  and $\hat{\epsilon}_e$ via Eq.\eqref{scala}. 
            \STATE $\epsilon_t =\mathcal{M} \odot \hat{\epsilon}_g + (1-\mathcal{M}) \odot \hat{\epsilon}_e$ 
            \STATE $x_{t-1} \leftarrow \epsilon_t$ via Eq.\eqref{xt_1}
		\ENDFOR
		\RETURN $x_0$.
	\end{algorithmic}
\end{algorithm}

\subsection{Saliency-aware Blending}
The above discussions are all about a single diffusion model, and now let us move to the next stage, i.e., obtaining a blending mask based on two models' salience maps.

Given a general model and an expert model, we can obtain the corresponding salience maps by Eqn.~\ref{saliency}, which are denoted as $\Lambda^{g}$ and $\Lambda^{e}$, respectively. We first normalize these salience maps by $softmax$ function:

\begin{equation}\label{lambda_}
\begin{aligned}
    &\Lambda^{g'} = softmax(k^g*\Lambda^g) \\
    &\Lambda^{e'} = softmax(k^e*\Lambda^e),
\end{aligned}
\end{equation}
where $k^g$ and $k^e$ are hyper-parameters, \textit{i.e.}, temperature of the $softmax$. Note that the $softmax$ here ensures the sum of each salience map to be a constant (\textit{i.e.}, 1), that means each model must focus on some regions instead of have high response on everywhere. When we integrate a general model and an expert model, the salience map of the general model tend to cover multiple objects to compose the whole scene, while the expert model tend to have higher response on a specific object. Thus we obtain a blending mask $\mathcal{M}$ by comparing these two salience maps:

\begin{equation}
\mathcal{M}=\arg\max(\Lambda^{g'},\Lambda^{e'})
\end{equation}

The saliency-aware mask is an effective guidance to perform noise blending, which consists of binary values of 0 and 1, corresponding to the noise $\epsilon_g$ and $\epsilon_e$ respectively. We can obtain the fused noise as follows:

\begin{equation}
     \hat\epsilon =\mathcal{M} \odot \hat{\epsilon}_g + (1-\mathcal{M}) \odot \hat{\epsilon}_e,
\end{equation}
where $\odot$ denotes Hadamard (element-wise) Product, and we omit $t$ for simplicity. Algorithm~\ref{HPLA} summarizes the process of the saliency-aware noise blending algorithm.

\textit{Additional explanations on the three applications.} In the three applications (\textit{i.e.}, a general model + a fine-grained model, a DreamBooth model, and a cartoon model) of this work, the expert model of the former two focuses on a specific object, while the last one contributes to the global structure of the generated image. In the last application, the cartoon model tends to focus on low-frequency structure and the general model focuses on high-frequency details. Thus the blended mask is not object-level, and we remove the blur operation in Eqn.~\ref{saliency} to facilitate such blending.  

\textit{Clarification of technique novelty.} The overall process of this algorithm is quite simple, yet it is non-trivial by solving two challenges. Firstly, we leverage the classifier-free guidance to automatically identify each model's areas of expertise. The introduction of the hyper-parameter k provides improved controllability for blending the two sources of images, thereby enabling greater creativity and flexibility in image generation based on SNB. Secondly, the task of obtaining saliency response values that are closely linked to prompt content is highly non-trivial. In each sampling step, the two models take as input the blended $x_t$, enabling automatic semantic alignment of the two models' noise space. We believe our exploration would contribute to the community and benefit the leveraging of pre-trained diffusion models.

\section{Experiments}
\label{exp}

\begin{figure*}[t]  
\begin{center}
   \includegraphics[width=0.9\linewidth]{ 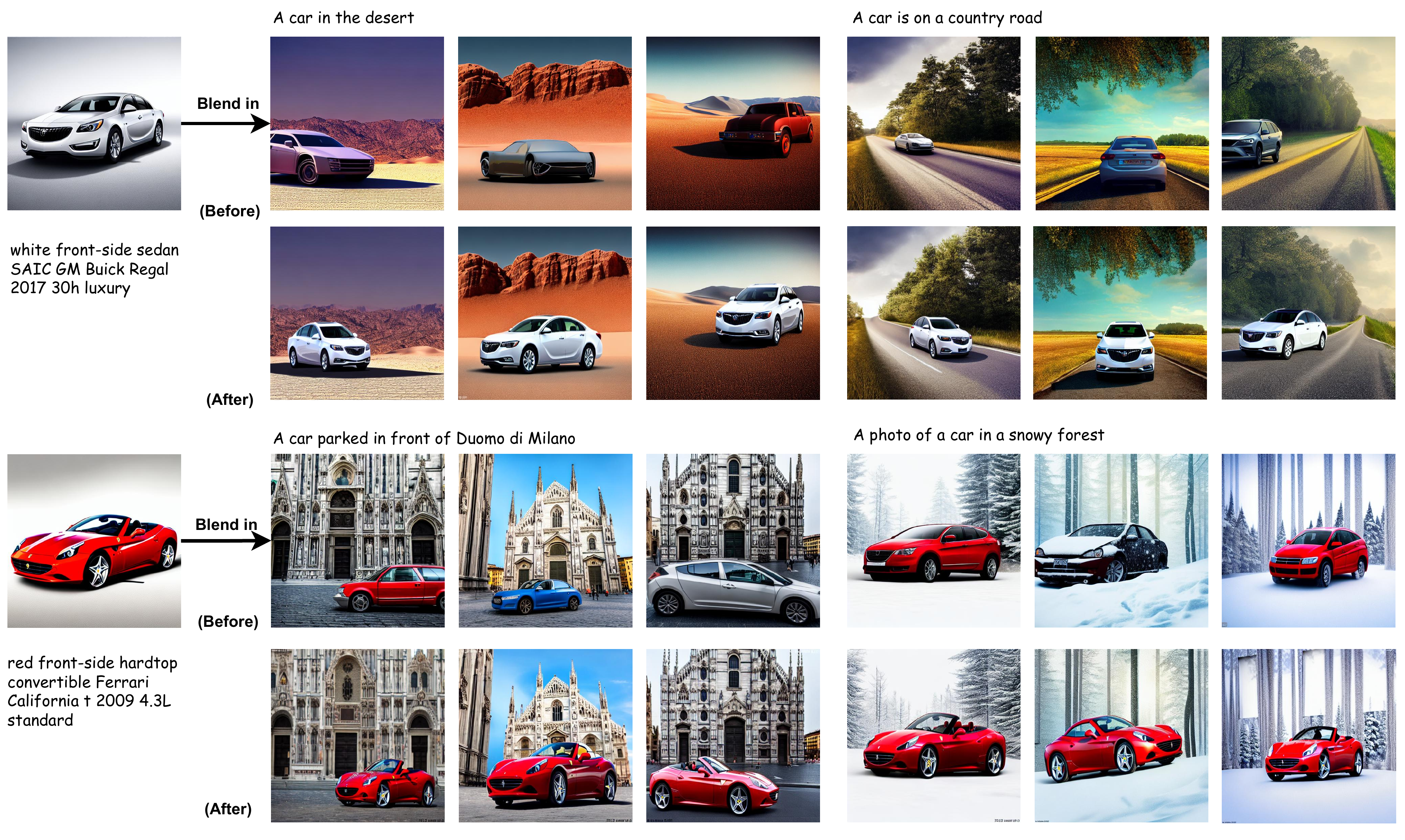}
\end{center}
   \caption{Visualization results of Fine-grained Fusion. Our method enables fine-grained generation with complex scenes.}
\label{fig:app1}
\end{figure*}

\begin{figure*}[t]  
\begin{center}
   \includegraphics[width=0.9\linewidth]{ 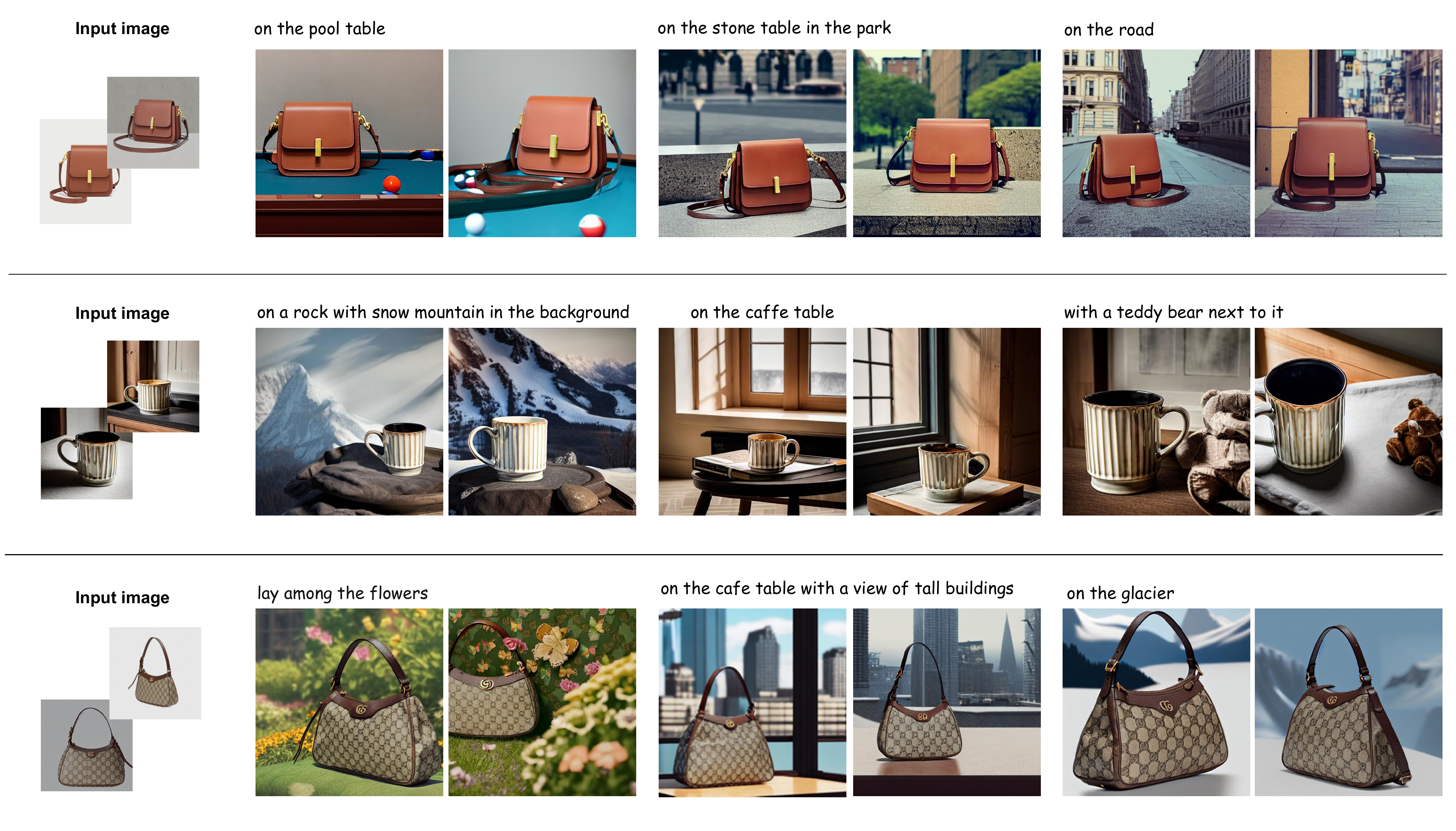}
\end{center}
   \caption{Visualization results of Recontextualization. Our method can recontextualize specific objects in complex scenes with well-preserved details.}
\label{fig:app2}
\end{figure*}

\begin{figure*}[t] 
\begin{center}
   \includegraphics[width=0.9\linewidth]{ 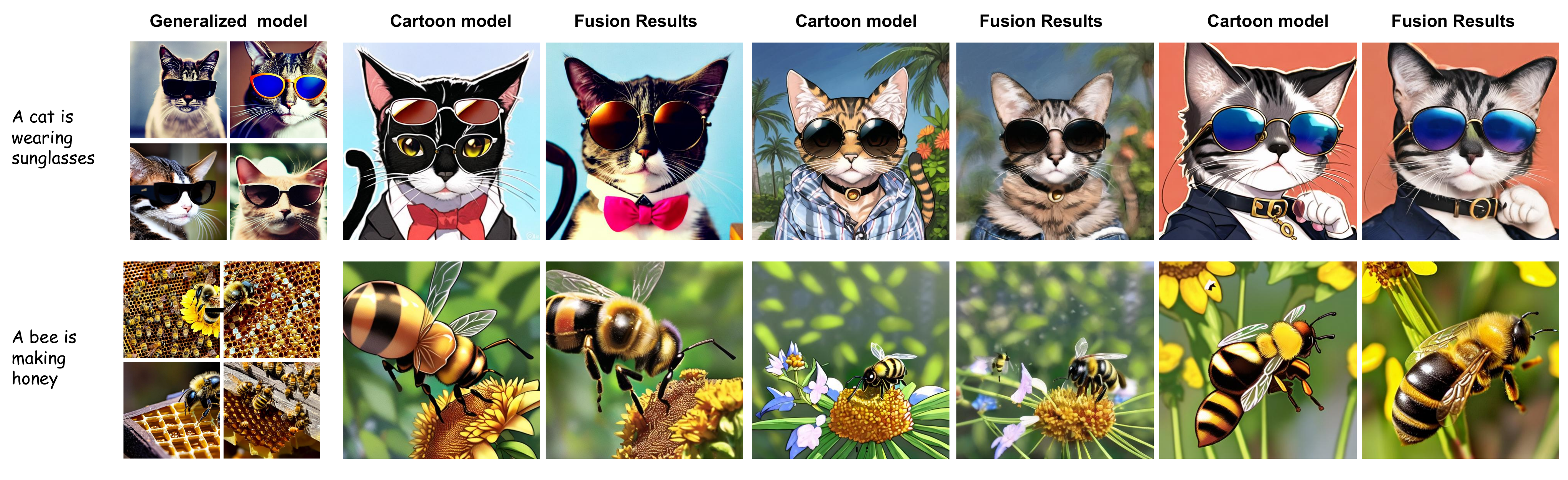}
\end{center}
\vspace{-5mm}
   \caption{Visualization results of Cross-domain Fusion. Our method can combine the creative advantages of the cartoon model to generate complex scenes and the photorealistic fidelity of the general model.}
   
\label{fig:cartoon}
\end{figure*}

\subsection{Applications}

In order to evaluate the effectiveness of our proposed method, we  conduct experiments on three challenging applications. 1) Fine-grained Fusion, \textit{i.e.}, fusing a general and a fine-grained model to achieve fine-grained generation with complex scenes. 2) Recontextualization, \textit{i.e.}, fusing a general and a DreamBooth~\cite{ruiz2022dreambooth} model allows for the recontextualization of specific objects with well-preserved details.  3) Cross-domain Fusion, \textit{i.e.}, fusing a general and a cartoon model to combine the creative advantages of the cartoon model to generate complex scenes and the photorealistic fidelity of the general model. These experiments will allow us to evaluate the performance of our method across a range of scenarios and provide valuable insights into the strengths and limitations of our approach.

\subsubsection{Application 1: Fine-grained Fusion}

The stable diffusion model~\cite{rombach2022high} trained on large-scale multimodal datasets like LAION 5B~\cite{schuhmann2022laion} has shown impressive performance on general text-to-image synthesis. In our experiments, we use the publicly released stable diffusion v1-4. Meanwhile, fine-grained car models trained on (an extension of) CompCars~\cite{yang2015large} can generate fine-grained car images with specific colors, viewpoints, types, brands, and models.

For instance, a scene description like ``a photo of a car in a snowy forest" can be fed into the general model, resulting in a corresponding image as shown in the first row of the right column in the second group of Figure \ref{fig:app1}.  Similarly, a specific car description like ``red hatchback Audi (imported) Audi A1 2010 e-tron" can be fed into the car model, generating an Audi car that matches the prompt. To fuse the noise of the two models during the denoising sampling process, we propose the SNB method, which can produce a fused image of a red Audi hatchback car in a snowy forest.  The fusion results of the general model and the fine-grained car model are illustrated in Figure \ref{fig:app1}, showing the ability to replace the car in the scene with a specific one while retaining the original scene unchanged. Notably, our SNB does not require additional annotations to specify the car's position in the original image, enabling automatic semantic alignment.

\subsubsection{Application 2: Recontextualization}

Recontextualization, or named personalizing text-to-image generation, is proposed in previous works ~\cite{gal2022image, ruiz2022dreambooth}, which aims to generate a creative scene for a specific object/concept. DreamBooth~\cite{ruiz2022dreambooth} proposes to fine-tune a diffusion model on several given images together with a placeholder word to enable the model to generate a specific object/concept. In this application, we integrate a general and a DreamBooth~\cite{ruiz2022dreambooth} model to allow for the recontextualization of specific objects with well-preserved details.

Specifically, we first fine-tune the general model using multiple images of the target object. Thus the fine-tuned model can represent the specific object with the placeholder ``[ ]" in the prompt. Next, we get through a sentence that describes a complex scene and ``a photo of a [ ] $<class>$'' into SNB to integrate the general model and the DreamBooth model. Results can be found in Figure ~\ref{fig:app2}. It can be observed that our method can put a specific object with rich details into a complex scene.

\subsubsection{Application 3: Cross-domain Fusion} 
Different models have distinct advantages when it comes to generating images with specific styles. For example, cartoon models are particularly skilled at creative composition and can generate scenes that are rarely observed in real-world scenarios. When we strive to generate images with unique and imaginative compositions, we can integrate a cartoon model to generate a distinctive scene and a general model to make the content more realistic. 

In this application, the same textual description of a scene is fed into both the general and the cartoon models. This allows us to generate two sets of noise that correspond to the same content but different styles during the sampling process. By fusing these two sets of noise, SNB generates images that exhibit both a creative composition and realistic content. As shown in Figure~\ref{fig:cartoon}, our proposed SNB provides a powerful tool for achieving the balance of creative and realistic, and we believe that such a tool has the potential to benefit a wide range of applications in various fields, such as art and AI-aided design.

\subsection{Comparisons and Ablations}
\begin{figure*}[!h]  
\begin{center}
      \includegraphics[width=0.9\linewidth]{ 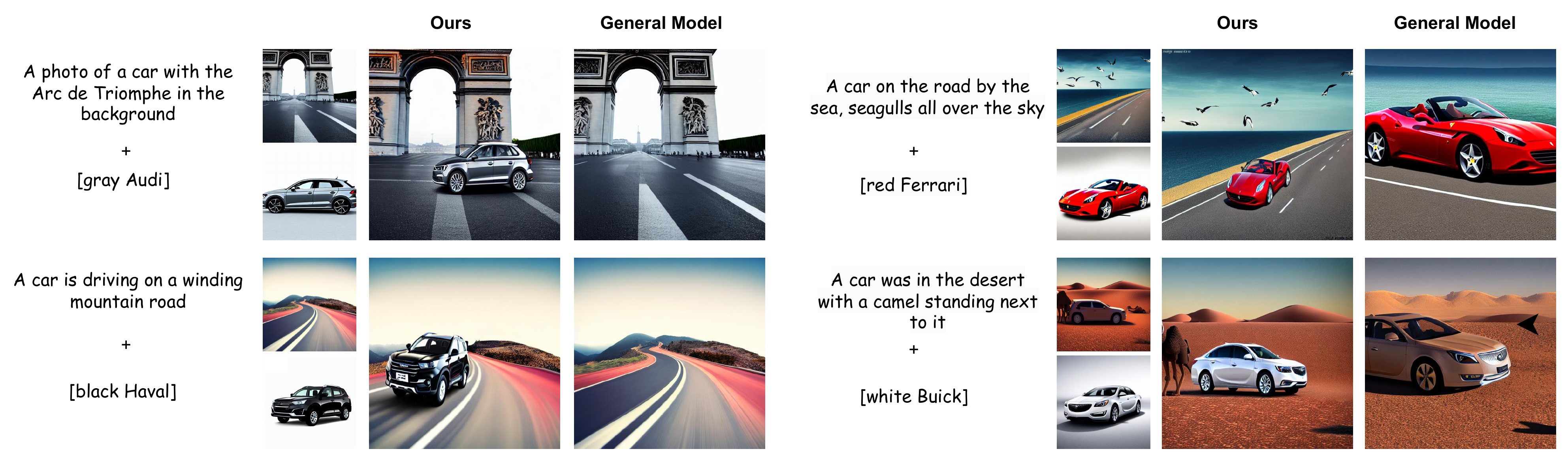}
\end{center}
   \caption{Compared to the general model. The general model fails to precisely generate fine-grained objects and tends to miss some objects when the prompt is full of details.}
\label{fig:6}
\end{figure*}

\begin{figure*}[t] 
\begin{center}
   \includegraphics[width=0.9\linewidth]{ 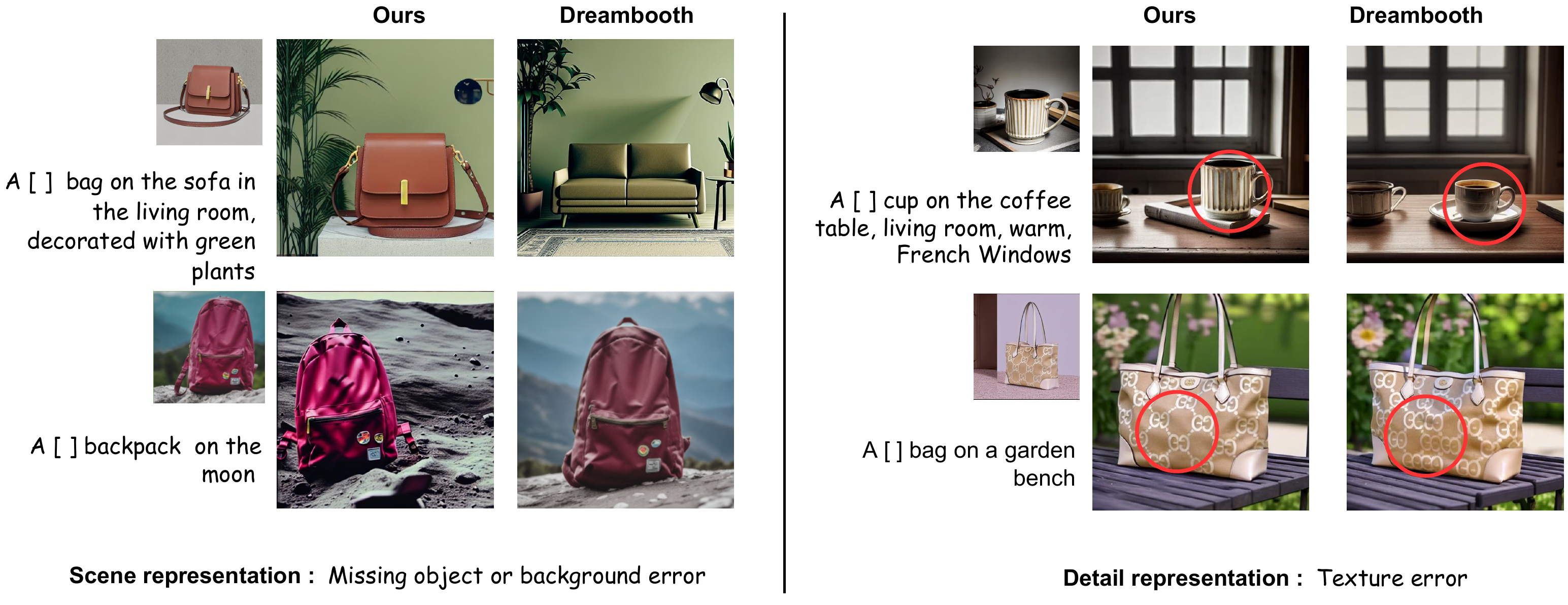}
\end{center}
   \caption{Compared to DreamBooth~\cite{ruiz2022dreambooth}. In situations where scene descriptions are complex, Dreambooth runs the risk of losing the object and faces difficulties in processing creative scene compositions like `` a [] backpack on the moon". In addition, Dreambooth struggles with preserving the details of the given objects, especially for complex and lengthy prompts.}
\label{fig:dreambooth}
\end{figure*}

\begin{figure*}[t]  
\begin{center}
   \includegraphics[width=0.8\linewidth]{ 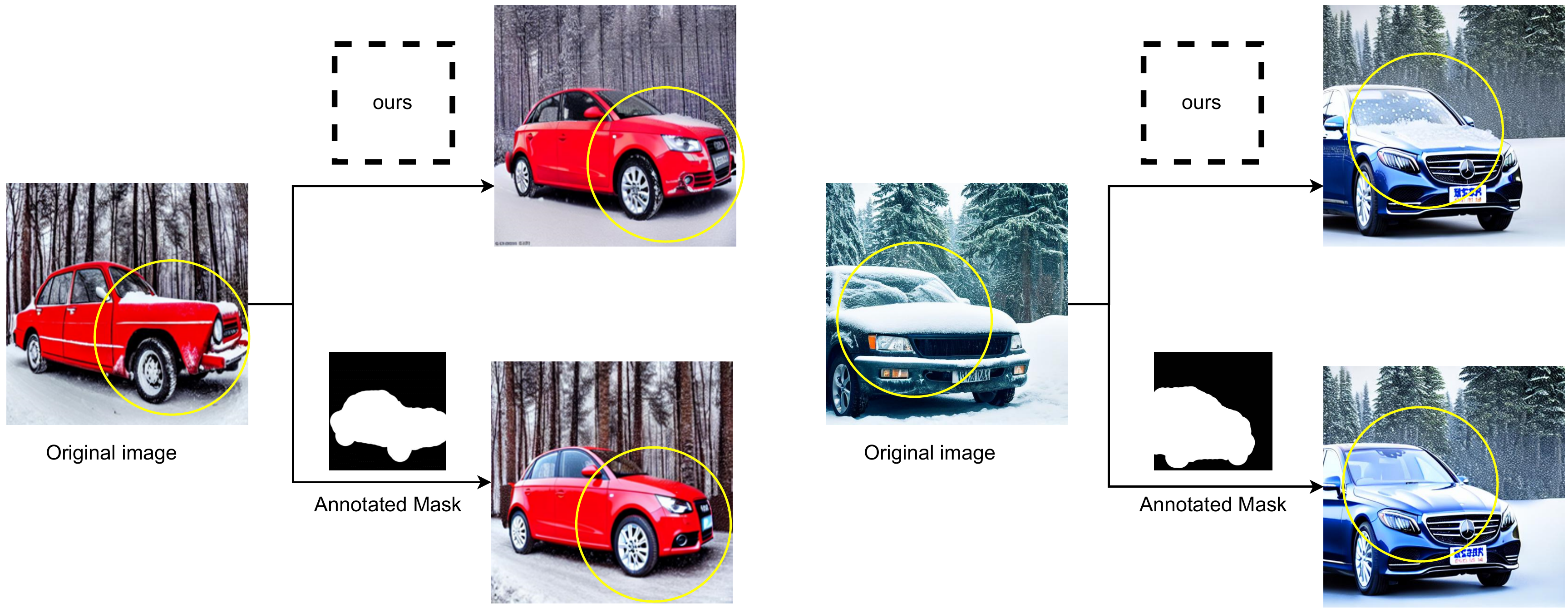}
\end{center}
   \caption{Compared to annotated masks. Our saliency-aware mask achieves better composing performance, e.g., generating snow on the car.}
\label{fig:mask}
\end{figure*}

\begin{figure*}[t]  
\begin{center}
   \includegraphics[width=0.87\linewidth]{ 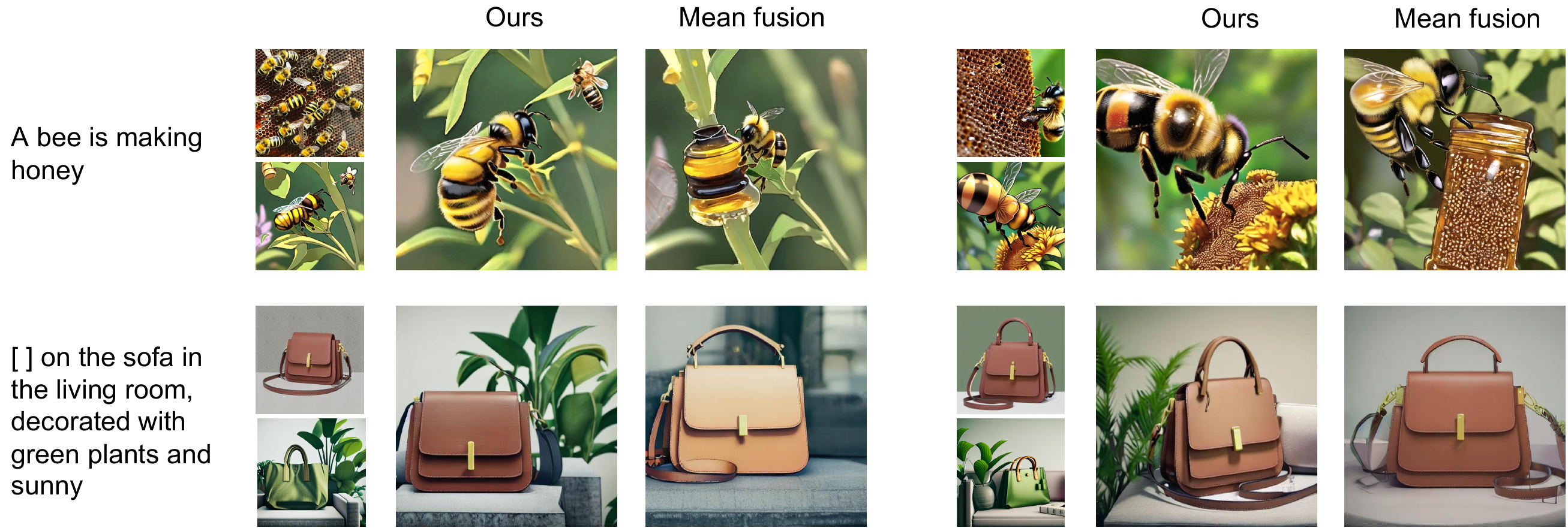}
\end{center}
   \caption{Compared to weighted sum. Directly averaging the two noises cannot well preserve each model's strengths and leads to cluttered image content or even the loss of crucial scene elements.}
\label{fig:mean}
\end{figure*}

\subsubsection{Compared to the General Model}
The stable diffusion model is a widely used and versatile generative model in computer vision that can generate a diverse range of images based on prompts. However, when the prompts contain multiple content subjects, the general model's generation performance can become challenging, especially when the combination of these subjects is rare in real-world scenes. This limitation can result in missing subjects in the generated images, as observed in Application 1, where the general model fails to capture both the car and the seagulls in the same scene as show in Figure~\ref{fig:6}. A similar issue is observed in Application 3, where the general model cannot produce scenes that are uncommon in real-life, such as a lion wearing a crown (Figure~\ref{fig:teaser} (c)). In addition, the general model has difficulty understanding the semantics of certain prompts, as observed in the case of ``a bee is making honey" in Figure~\ref{fig:cartoon}, where the model only combines the bee and honey in the same scene. 

The proposed SNB can address these limitations, which integrates a specialized model, such as a model that focuses on cars in Application 1, or a cartoon model in Application 3, to improve the accuracy and realism of image generation. By leveraging the strengths of both the general and specialized models, our proposed approach can generate visually appealing and semantically rich images. This approach has the potential to advance the field of computer vision by enabling more sophisticated and realistic image generation.

\subsubsection{Compared to Dreambooth}
Dreambooth \cite{ruiz2022dreambooth} presents an approach for synthesizing novel renditions of a given subject using a few images of the subject and the guidance of a text prompt, \textit{i.e.}, recontextualization. However, it is challenging for Dreambooth to handle certain creative scene compositions, such as ``a [] backpack on the moon."  Our empirical investigation reveals that Dreambooth struggles with integrating specific subjects into the scene, particularly for complex and lengthy prompts.  As illustrated in Figure~\ref{fig:dreambooth}, Dreambooth frequently produces images that either lack or exhibit inconsistencies in terms of specific subjects in such scenarios.  In contrast, our approach excels at integrating a specific subject into the scene while maintaining fidelity to the prompt.  Although Dreambooth can generate contextually relevant images for brief prompts like ``a [] bag on a garden bench." it may not accurately capture the finer details of the subject, such as the graphic texture of the bag.  In contrast, our approach not only generates images that adhere to the prompt but also reproduces subject details with greater accuracy.

\subsubsection{Compared to Annotated Masks}
Blended-Diffusion~\cite{avrahami2022blended} firstly propose to blend the noisy image in each sampling step based on a given mask for image editing. Such a method can neither handle applications like cross-domain fusion nor perform well on mixed object and scene content, such as ``a car in a snowy forest." As depicted in Figure \ref{fig:mask}, a comparison between our method and a mask-based approach demonstrates that our SNB method accurately preserves detailed subject features influenced by the scene, like the snow on the car's hood and wheels. In contrast, the mask-based method replaces the entire car area, resulting in a loss of subject details within the scene. Overall, our SNB yields more natural and realistic results when editing or replacing scene content, particularly in scenarios with mixed object and scene content.

\subsubsection{Compared to Weighted Sum}
The success of SNB hinges upon its ability to create a high-quality mask based on the noise generated by the two models, which is then used to determine content coverage and retention. As shown in Figure \ref{fig:mean}, the experimental results of SNB and the fusion method that directly averages the two noises differ significantly.  When the weighted sum of two noises is used, the resulting image content appears cluttered and lacks the desired semantic alignment. In the task of recontextualization, there's a risk of losing important scene content due to inaccurate masking. In contrast, our SNB method outperforms the direct averaging method by achieving precise semantic alignment, resulting in more accurate and realistic image content.

\section{Conclusion}
\label{con}
In this work, we study the problem of integrating pre-trained text-guided diffusion models to achieve more controllable generation. We propose a simple yet effective Saliency-aware Noise Blending (SNB) to preserve the strengths of each individual model. Extensive experiments on three challenging applications (\textit{i.e.}, a general model + a cartoon model, a fine-grained car model, and a DreamBooth model) show that SNB can significantly empower pre-trained diffusion models. With the rapid development of pre-trained large generative models, we believe our work is of great value. In the future, we will keep on studying the integration of different large generative models and   extend our method to more general settings.

{\small
\bibliographystyle{ieee_fullname}
\bibliography{egbib}
}

\clearpage
\section*{Appendix}


\begin{figure*}[b]
\begin{center}
   \includegraphics[width=1\linewidth]{ 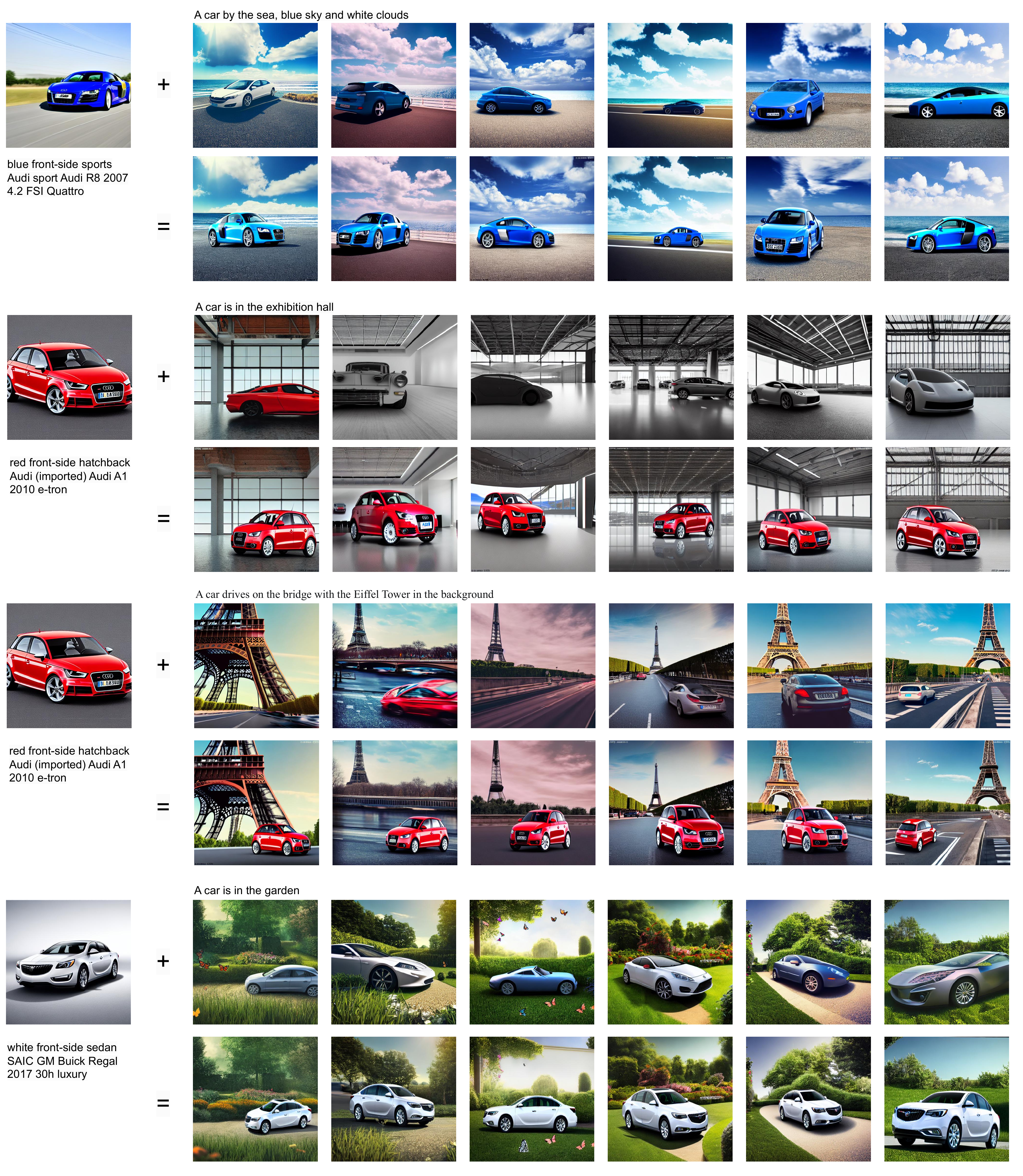}
\end{center}
   \caption{Supplementary experimental results for Application 1.}
\label{fig:10}
\end{figure*}

\begin{figure*}[t]
\begin{center}
   \includegraphics[width=1\linewidth]{ 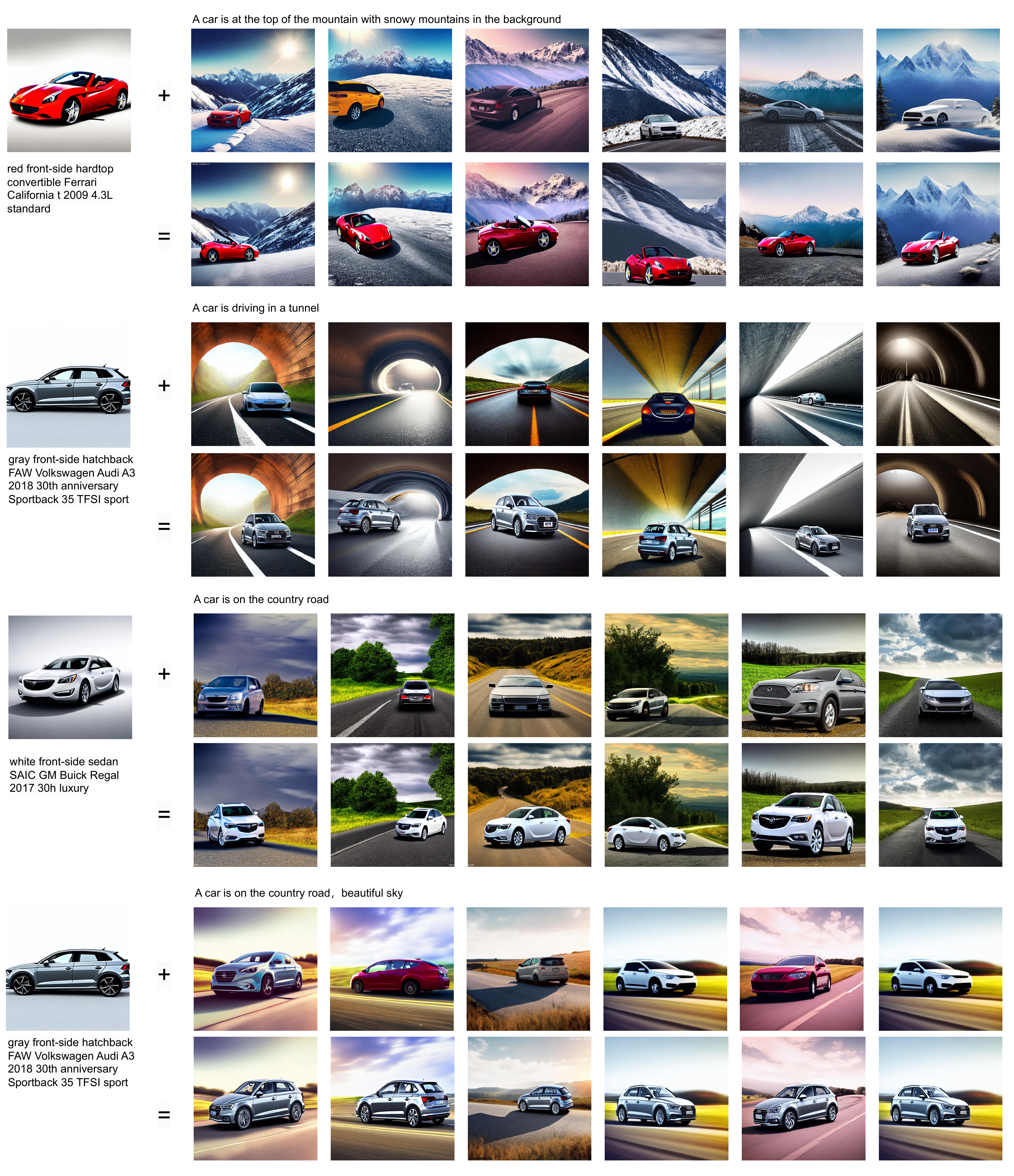}
\end{center}
   \caption{Supplementary experimental results for Application 1.}
\label{fig:11}
\end{figure*}

\begin{figure*}[t] 
\begin{center}
   \includegraphics[width=1\linewidth]{ 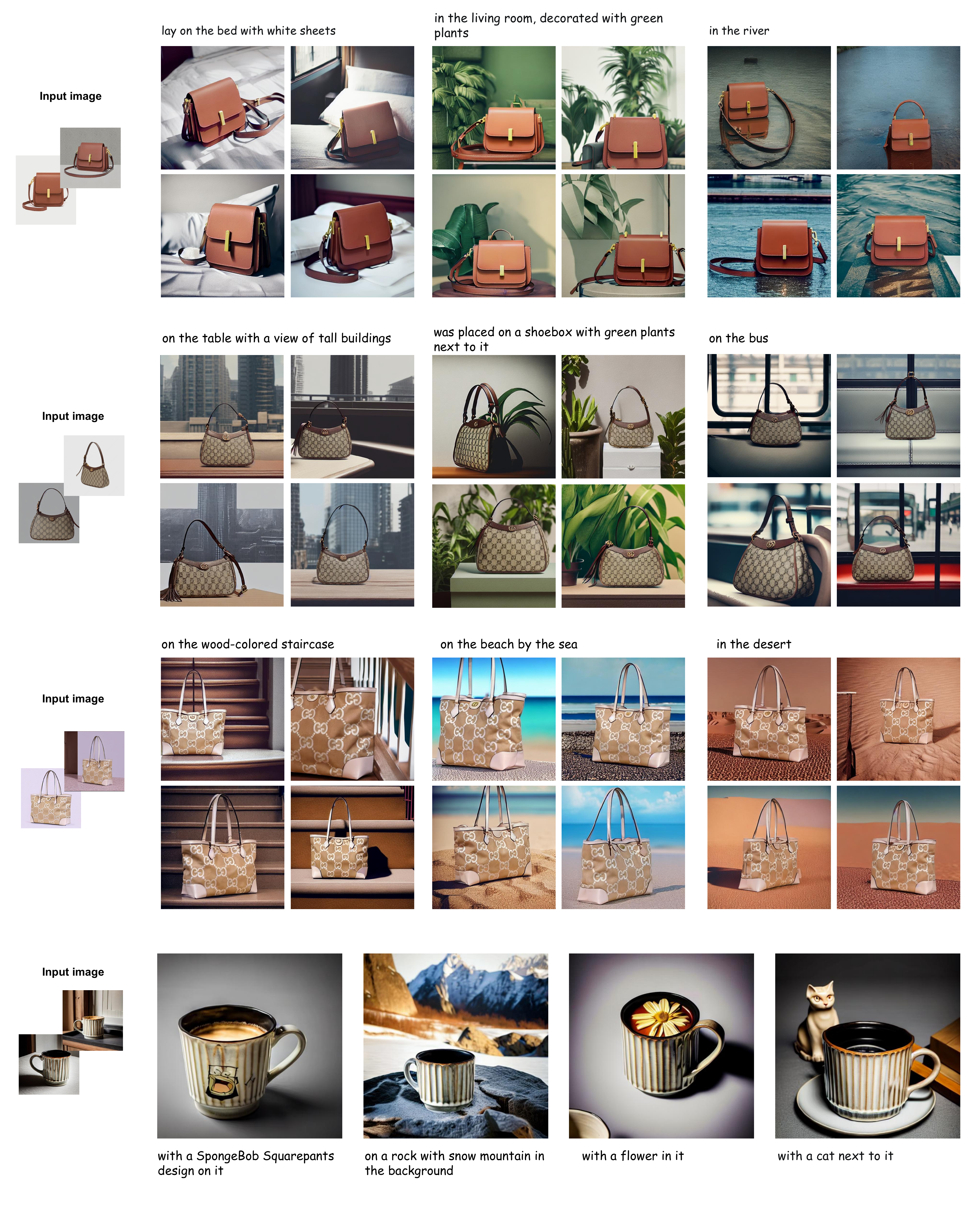}
\end{center}
\vspace{-5mm}
   \caption{Supplementary experimental results for Application 2.}
\label{fig:12}
\end{figure*}

\begin{figure*}[t] 
\begin{center}
   \includegraphics[width=1\linewidth]{ 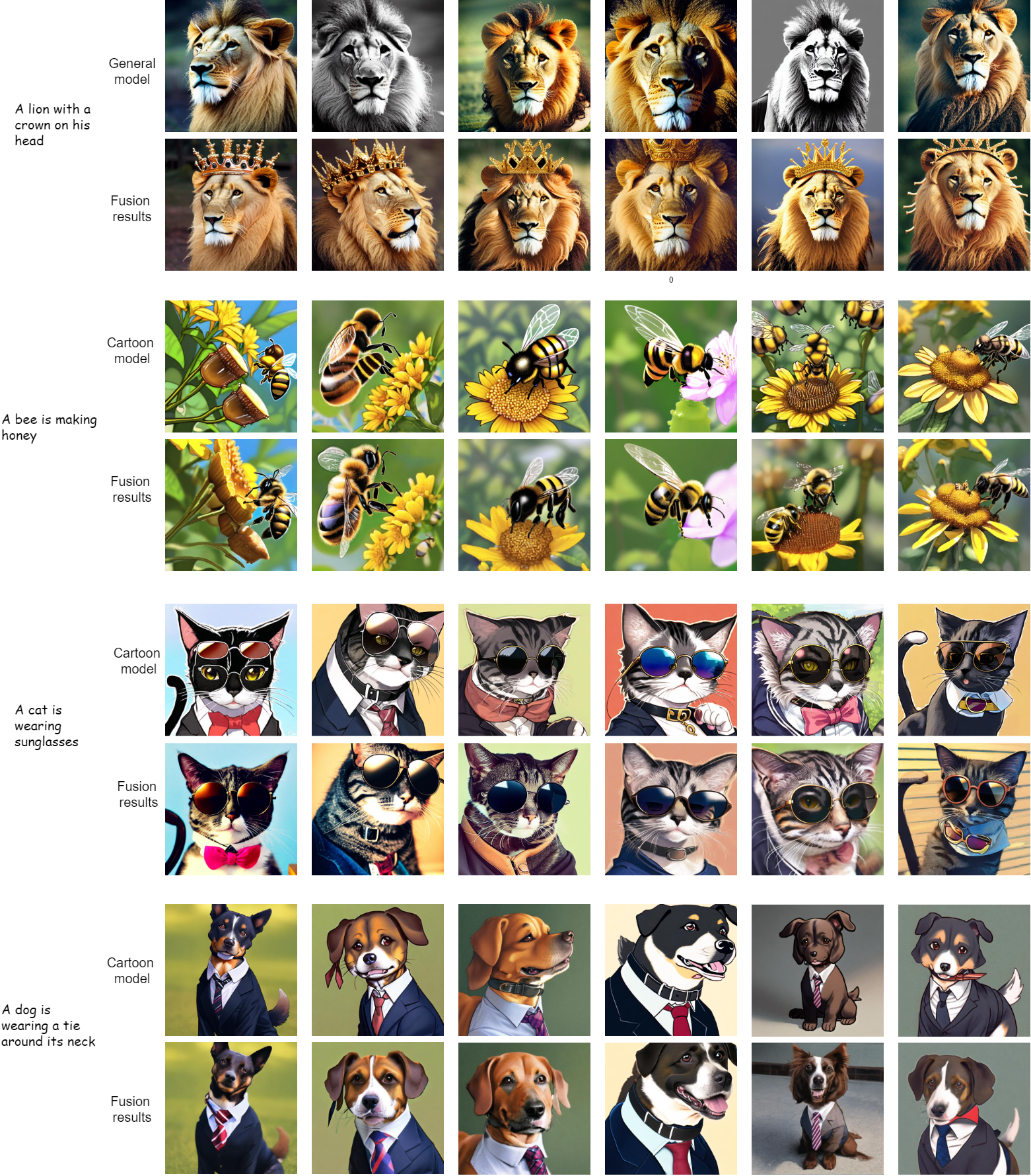}
\end{center}
   \caption{Supplementary experimental results for Application 3.}
\label{fig:13}
\end{figure*}

\begin{figure*}[t] 
\begin{center}
   \includegraphics[width=1\linewidth]{  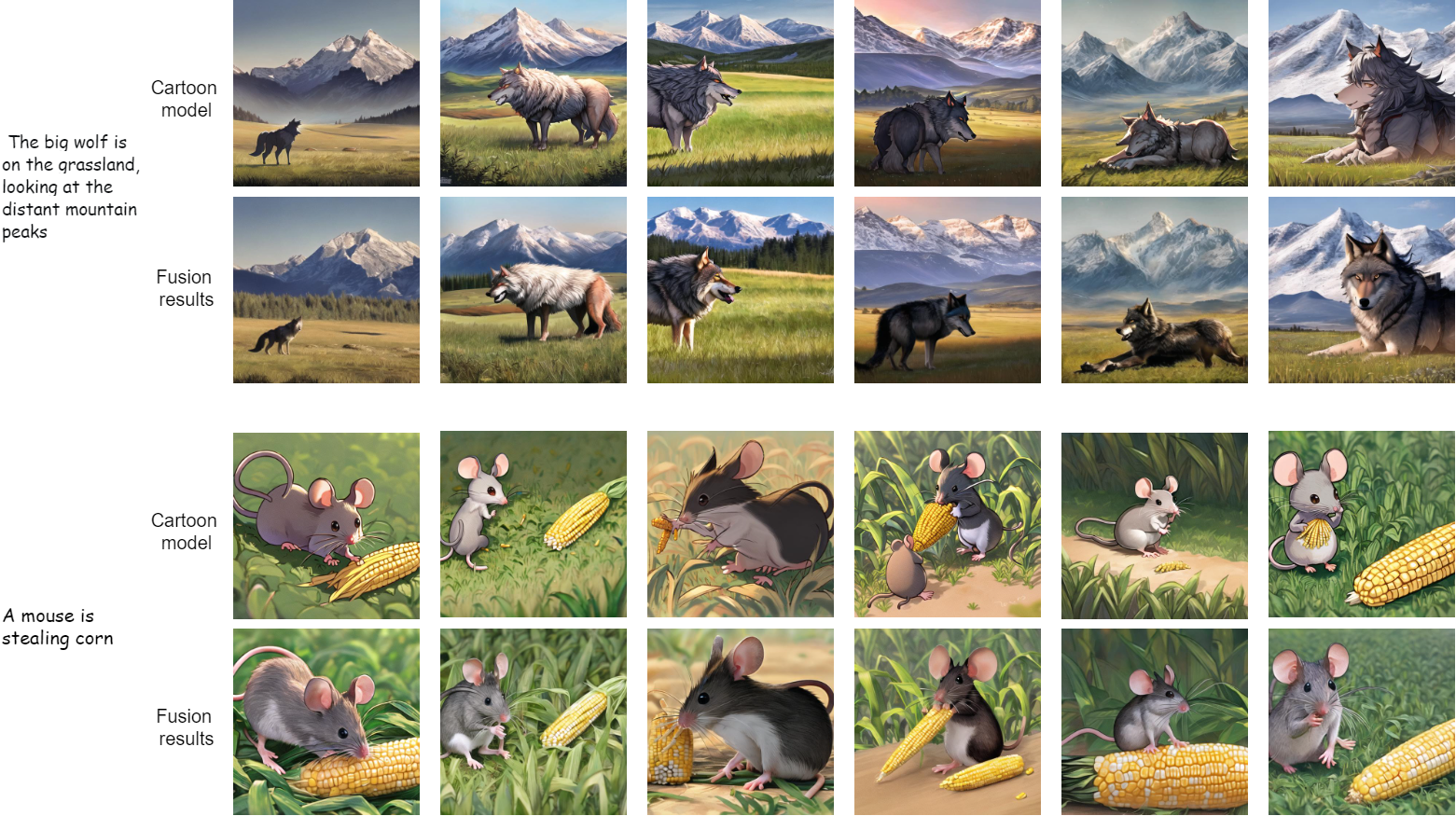}
\end{center}
   \caption{Supplementary experimental results for Application 3.}
\label{fig:13}
\end{figure*}

\end{document}